\documentclass{article}

\usepackage{microtype}
\usepackage{graphicx}
\usepackage{booktabs} 

\usepackage{hyperref}

\usepackage[accepted]{icml2023}

\usepackage{amsmath}
\usepackage{amssymb}
\usepackage{mathtools}
\usepackage{amsthm}

\usepackage[capitalize,noabbrev]{cleveref}

\usepackage{url}

\usepackage[utf8]{inputenc} 
\usepackage[T1]{fontenc}    
\usepackage{url}            
\usepackage{booktabs}       
\usepackage{amsfonts}       
\usepackage{nicefrac}       
\usepackage{microtype}      
\usepackage{xcolor}         
\usepackage{soul}
\usepackage{array}
\usepackage{graphicx}
\usepackage{subfigure}
\usepackage{amsmath, bm}
\usepackage{dsfont}
\usepackage{amssymb}
\usepackage{multirow}

\usepackage[vlined,linesnumbered,ruled,algo2e]{algorithm2e}
\usepackage[linesnumbered,ruled,vlined]{algorithm2e}
\usepackage[flushleft]{threeparttable}

\usepackage{varwidth}
\usepackage{pifont}
\usepackage{makecell}
\usepackage{wrapfig}
\usepackage{enumitem}
\usepackage{caption}
\usepackage{dsfont}
\usepackage{soul}
\usepackage{varwidth}
\usepackage{pifont}
\usepackage{makecell}
\usepackage{enumitem}

\usepackage[hang,flushmargin]{footmisc}

\usepackage{xspace}

\usepackage{microtype}
\usepackage{graphicx}
\usepackage{booktabs} 
\usepackage{xspace}
\usepackage{colortbl}
\usepackage{array,multirow,graphicx}
\usepackage{makecell}
\usepackage{textcomp}
\usepackage{multirow}

\DeclareMathOperator*{\minimize}{min}

\newcommand{\glam}{GLaM\xspace}

\theoremstyle{plain}

\theoremstyle{definition}

\theoremstyle{remark}

\usepackage[textsize=tiny]{todonotes}

\icmltitlerunning{Lifelong Language Pretraining with Distribution-Specialized Experts}

\begin{document}

\twocolumn[
\icmltitle{Lifelong Language Pretraining with Distribution-Specialized Experts}



\icmlsetsymbol{equal}{*}


\begin{icmlauthorlist}
\icmlauthor{Wuyang Chen}{equal,ut}
\icmlauthor{Yanqi Zhou}{google}
\icmlauthor{Nan Du}{google}
\icmlauthor{Yanping Huang}{google}
\icmlauthor{James Laudon}{google}
\icmlauthor{Zhifeng Chen}{google}
\icmlauthor{Claire Cui}{google}
\end{icmlauthorlist}

\icmlaffiliation{ut}{The University of Texas at Austin}
\icmlaffiliation{google}{Google}

\icmlcorrespondingauthor{Yanqi Zhou}{yanqiz@google.com}
\icmlcorrespondingauthor{Nan Du}{dunan@google.com}



\icmlkeywords{Machine Learning, ICML}

\vskip 0.3in
]



\printAffiliationsAndNotice{\icmlEqualContribution} 

\begin{abstract}
Pretraining on a large-scale corpus has become a standard method to build general language models (LMs). Adapting a model to new data distributions targeting different downstream tasks poses significant challenges.
Naive fine-tuning may incur catastrophic forgetting when the over-parameterized LMs overfit the new data but fail to preserve the pretrained features.
Lifelong learning (LLL) aims to enable information systems to learn from a continuous data stream across time. However, most prior work modifies the training recipe assuming a static fixed network architecture. We find that additional model capacity and proper regularization are key elements to achieving strong LLL performance. Thus, we propose Lifelong-MoE, an extensible MoE (Mixture-of-Experts) architecture that dynamically adds model capacity via adding experts with regularized pretraining.
Our results show that by only introducing a limited number of extra experts while keeping the computation cost constant, our model can steadily adapt to data distribution shifts while preserving the previous knowledge.
Compared to existing lifelong learning approaches,
Lifelong-MoE achieves better few-shot performance on 19 downstream NLP tasks.
\end{abstract}

\section{Introduction}

\begin{figure}[t!]
\includegraphics[scale=0.56]{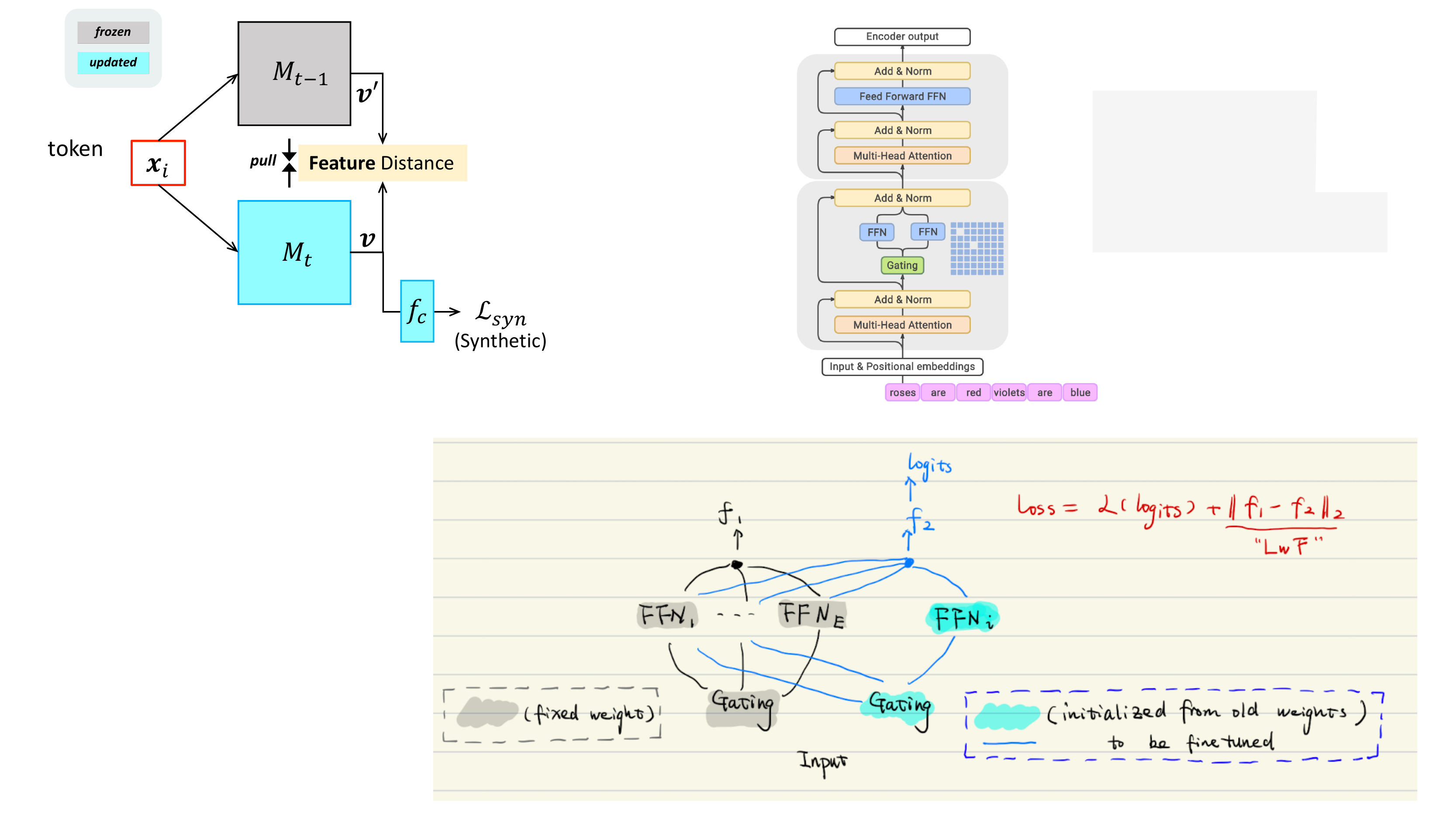}
\centering
\captionsetup{font=small}
\caption{Overview of our Lifelong-MoE method: 1) During pretraining, the expanded experts (and gatings) are specialized for each data distribution; 2) We freeze the pretrained old experts and gatings; 3) We further introduce regularizations to the MoE to avoid the catastrophic forgetting.}
\label{fig:teaser}
\vspace{-1em}
\end{figure}

Language models (LMs),
from word embeddings/vectors~\citep{mikolov2013efficient}, to recurrent neural networks~\citep{sutskever2014sequence}, and to the latest self-attention-based Transformer networks~\citep{vaswani2017attention},
play increasingly important roles in natural language processing (NLP) tasks, including both language generation and language understanding.
Recent works on scaling up both pretraining data and large models~\citep{shazeer2017outrageously,gpipe19,kaplan2020scaling} enable the inference on complicated NLP tasks with much less data, and fewer or even no additional label for downstream tasks.
For example, BERT~\citep{xu2019bert} and GPT-3~\citep{NEURIPS2020_gpt3} demonstrate that for few-shot or even zero-shot generalization on downstream corpus, current LMs only require very few labeled examples to achieve good generalization on unseen tasks.
More recently, GLaM~\citep{du2022glam} proposes using a sparsely activated mixture-of-experts architecture to scale the model capacity while incurring substantially less training cost compared to dense variants.

Pretraining large language models (LMs) has become the \textit{de facto} standard before adapting NLP models to downstream tasks.
This is extremely successful when the pretraining and downstream task are drawn from the same corpus distribution.
Most of time, benchmarking large LMs blindly assumes the existence of a static and well-balanced pretraining dataset.
While being accurate, the performance of large LMs on downstream tasks heavily relies on the high quality of large-scale pretraining, which is not always guaranteed in the wild for several reasons.
First, \ul{at the data level}, new language corpus (online forum conversations, new wikipedia pages, websites, book chapters, etc.) mostly emerges in a streaming online fashion. That means to keep our pretraining dataset up-to-date, new data distributions will be collected continuously, instead of being statically stored offline in batches. However, in real-world scenarios, sequentially pretraining LMs on new corpus samples with changing distributions will cause catastrophic forgetting on previously learned knowledge.
In addition, the collection and maintenance of such high-quality corpora is intensive in manual labor.
Second, \ul{at the optimization level}, pretraining a large LM is time and resource consuming, especially on an increasingly large pretraining corpus.
For example, pretraining a GPT-3 model with 280B language tokens requires over 500 TPU hours~\citep{du2022glam}. As the number of tokens in the pretraining set increases, the pretraining cost will keep rising.

In practice, it is highly preferred to continually pretrain LMs whenever a new corpus is collected, in order to reduce training codecodingst and enhance performance on previously out-of-domain data.
Despite its importance, the challenge of continually pretraining a large LM over online data streams is largely under-explored.
Lifelong learning (LLL) is a research topic on solving this data/task shifting issue. As opposed to computer vision or robotics, LLL is particularly challenging and nascent in the NLP domain~\citep{greco-etal-2019-psycholinguistics,sun2020LAMOL}, as natural language is compositional and context-dependent.
Prior works in LLL primarily focus on \textit{task-incremental} settings with \textit{boundary-aware} data streams.
Starting from the same pretrained checkpoint, these LLL methods are usually evaluated on a sequence of downstream tasks instead of pretraining data distributions~\citep{Aljundi2019OnlineCL}.
However, this task-level lifelong learning is not the most practically common setting in NLP, because: 1) pretaining is usually agnostic to downstream tasks; 2) as LMs are shown to be few-shot learners, a steam of downstream tasks will incur marginal or zero impact on the pretrained weights. Instead, any shift in pretraining data will pose real forgetting issues.

In this work, we target solving the \textit{data-level} lifelong pretraining with shifting distributions in NLP tasks, especially for large language models.
We aim at task-agnostic preservation of domain-specific knowledge from a sequence of online pretraining corpus distributions.
We start our method on top of the mixture-of-experts (MoE)~\cite{shazeer2017outrageously,lepikhin2020gshard,du2022glam}, with an intuition that MoE can increase its model capacity for fitting changing corpus distributions along the online data streams without incurring extra computation cost.
Our finding is that, by only introducing \ul{extra expert layers plus proper expert regularizations}, we can continuously pretrain a mixture-of-experts model on a sequence of data distributions without forgetting old knowledge, and achieve competitive or even better one-shot performance in downstream tasks.
The expanded experts will not increase the computation overhead, since they are always sparsely activated and only a fixed number of experts will be selected for each token.
Specifically, we show the benefits from three key lifelong learning strategies for MoE: 1) partially expanded experts and gating dimensions; 2) frozen old experts and gatings with only newly expanded ones to be optimized; 3) output-level regularization from previously pretrained knowledge.
With these three methods, we aim at creating a well-balanced trade-off between maintaining old knowledge and fitting new distributions.
Compared with the dense counterpart, our method can achieve competitive or even better decoding scores on one-shot downstream tasks, including the QA (question answering) task and the translation task.
Our contributions are summarized below:\vspace{-0.5em}
\begin{itemize}[leftmargin=*]
    \item We propose the first lifelong pretraining framework for large-scale mixture-of-experts (MoE) language models that is agnostic to downstream tasks.
    \item We progressively expand the number of experts to increase model capacity and fit new pretraining data distributions, and preserve old knowledge by freezing previously trained old experts and gatings.
    \item We carefully study the output-level regularization to allow dense layers in MoE to fit new data distribution without forgetting old distributions.
    \item We achieve state-of-the-art decoding scores on downstream one/zero-shot tasks, including the QA task, the translation task, and other language understanding tasks.
\end{itemize}

\section{Related Work}

\paragraph{Pretraining and Fine-tuning in Language Models}
Deep networks are shown to be powerful in many NLP tasks.
Works using recurrent networks such as RNNs and LSTMs~\citep{Mikolov2010rnnlm,sutskever2011_rnnlm} for word/sentence representations~\citep{NIPS2015_dai,kiros-skip-thought} show that language models can improve diverse NLP understanding tasks. More recently, self-attention and transformers~\citep{vaswani2017attention} demonstrate that larger models with unsupervised pretraining on unlabeled data can yield significant generalization on NLP problems~\citep{devlin2018bert,yang2019xlnet,liu2019roberta,clark2020electra}.
Abundant computation resources and corpus data makes the pretraining of increasingly large language models possible. These large language models leverage the scaling power of model size and the network's remarkable fitting capacity.
Transfer learning based on pretraining and finetuning~\citep{raffel2020exploring,houlsby2019parameterefficient} has been extensively studied and shows good performance on few-shot downstream tasks.
The problem of current pretraining and fine-tuning paradigm is that, updating the pretraining dataset will incur repeated heavy re-training cost.

\paragraph{Sparsely Gated Networks}
Despite the success of large and dense language models, training these networks requires significant amounts of computing resources.
To keep scaling up NLP models without incurring heavy computational cost, mixture-of-experts (MoE) is recently developed to enable sparse activations in dense layers, and demonstrates significant advantages.
For language modeling and machine translation, \citet{shazeer2017outrageously} shows that they can use a large number of parameters while only activating a small subset for each inference. The choice of dense layers to activate is controlled by a learnable gating function.
There is an increasing number of works on scaling sparsely activated MoE architectures~\citep{hestness2017deep,shazeer2018mesh,lepikhin2020gshard,kudugunta2021beyond}, including Switch-C~\citep{fedus2021switch} and GLaM~\citep{du2022glam}.
All these MoE efforts show greatly reduced training energy and computation cost, while still achieving better overall zero, one, and few-shot performance across diverse NLP tasks and domains~\cite{gururangan2021demix}.
In this work, we will show a further advantage of MoE: the expanded experts and gatings can enlarge the model capacity of multiple data distributions without introducing computation overhead. Besides, we only implicitly ``assign'' experts to different domains instead of any explicit conditions.

\paragraph{Continual Learning for NLP.}
In general, solutions proposed for lifelong learning can be classified into the following categories: i) replay based approaches~\citep{robins1995catastrophic,rebuffi2017icarl,shin2017continual,lopez2017gradient,chaudhry2018efficient}; ii) regularization based approaches~\citep{ewc2017,li2018learning}; iii) architecture based approaches~\citep{rusu2016progressive,yoon2017lifelong,Mallya2018PackNetAM,Wen2020BatchEnsemble}.
Recently, lifelong learning is drawing attention for NLP problems~\citep{wang2019sentence,biesialska2020continual,Sun2020LAMOLLM, huang2021continual,hussain2021towards,ahrens2021drill,clif,lin2022continual}.
A number of lifelong learning methods have also been proposed, including embedding aligned episodic memory replay~\citep{wang-etal-2019-sentence}; memory-based parameter adaptation with sparse experience replay (MbPA++)~\citep{dautume2019episodic}; language modeling for lifelong language learning~\citep{sun2019lamol}; and meta-learning with sparse experience replay~\citep{holla2020meta}.
The primary challenge to address in LLL literature is to overcome the catastrophic forgetting.
However, most works still focus on the traditional settings on sequential downstream tasks, ignoring the fact that pretrained large language models have the capability to quickly adapt to downstream tasks with only a few samples. This task-level lifelong learning is not directly beneficial to most of the real-world scenarios of deployed NLP models, as downstream tasks marginally update model parameters.
In contrast, we focus on continually pretraining language models on a steam of changing data distributions (i.e. the data-level lifelong pretraining). This setting is more close to practical scenarios to continually deploying and updating language models.

\begin{figure*}[h!]
\includegraphics[scale=0.4]{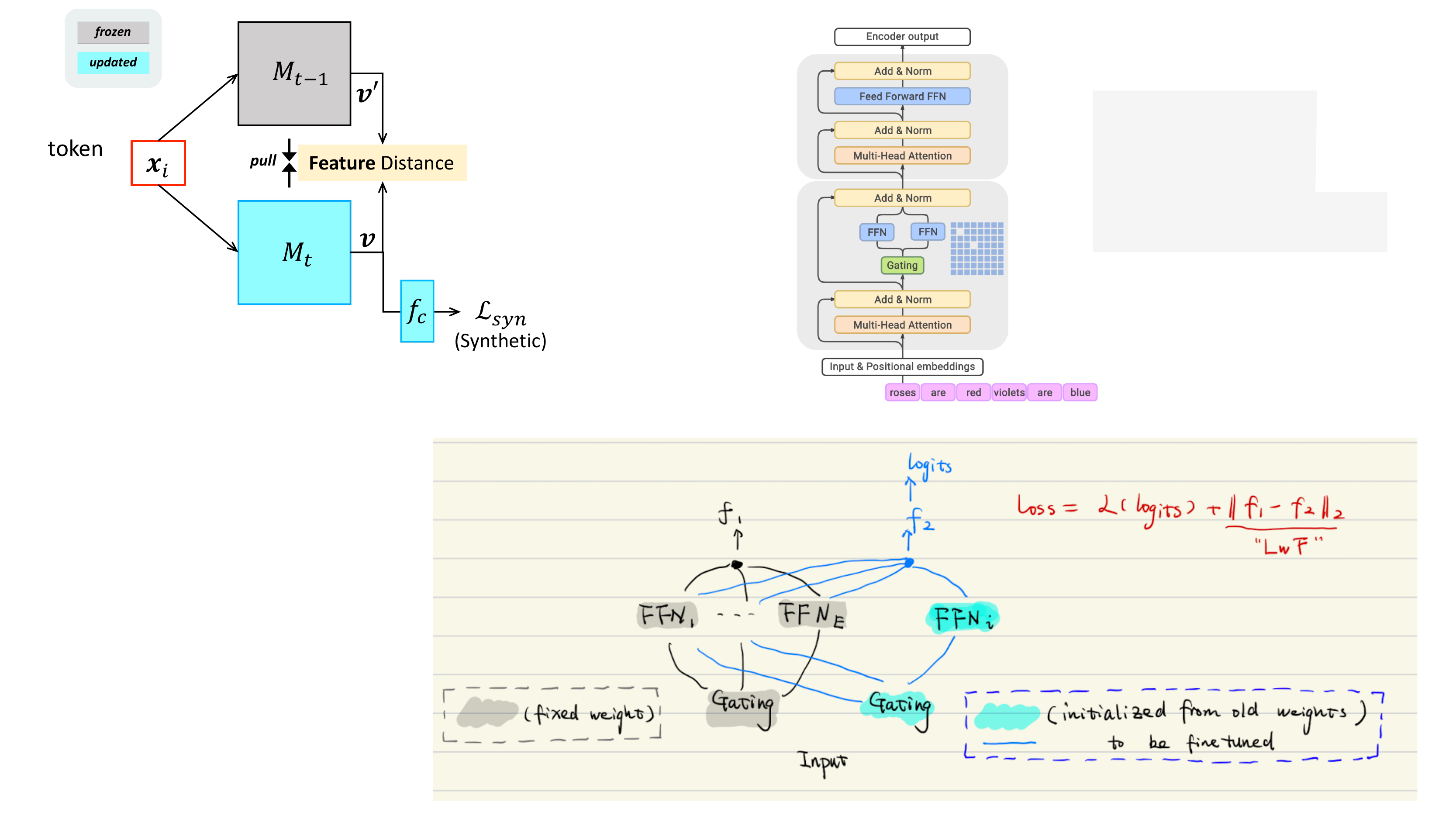}
\centering
\vspace{-0.5em}
\captionsetup{font=small}
\caption{Overview of our lifelong pretraining method for the MoE model ($\mathcal{M}$): 1) When pretraining on each data distribution ($\bm{x}^{(t)}$), we expand the number of experts and gatings (from $E^{(t-1)}$ to $E^{(t)}$) for larger model capacity; 2) We freeze the pretrained old experts and gatings; 3) We further regularize the MoE on the output level to avoid the catastrophic forgetting.
Embedding, dense, and attention layers (omitted in this figure) are shared across all data distributions. See details of our method in Section~\ref{sec:method} and pretraining settings in Section~\ref{sec:pretraining}. We omit the interleaving dense layers to make this figure simple and clear.}
\label{fig:framework}
\vspace{-1em}
\end{figure*}

\section{Pretraining MoE without Forgetting} \label{sec:method}

Experts and gatings play a vital role in determining MoE's capability of adapting to new data distributions. This motivates us to develop a lifelong pretraining method by only focusing on the customization of experts and gatings.
Our strategy is designed as follows: 1) to ensure enough capacity of the MoE whenever it fits a new data distribution, we will expand (and only expand) the number of experts and gating dimensions, keeping the network's depth and width unchanged; 2) to avoid the expanded MoE from overfitting the training data, we will introduce proper regularization on experts and gatings and encourage the preservation of previously learned knowledge.

\subsection{Model Architecture}
\label{sec:model}

We leverage \glam~\citep{du2022glam} as our base model,
a family of sparsely activated Mixture-of-Experts (MoE)~\citep{shazeer2017outrageously, fedus2021switch}.
We are motivated in solving the lifelong pretraining problem in NLP
by only introducing more parameters without introducing extra computation overhead (as we will always only use token-wise top-2 experts during both training and inference).

Based on the GShard Transformer~\citep{lepikhin2020gshard},
\glam replaces the feed-forward component of every other transformer layer with an MoE layer. Each MoE layer consists of a collection of independent feed-forward dense layers as the ``experts''. A gating function uses softmax to calculate a probability distribution to indicate the preference of the input token to each expert.
The dimension of gating's weight equals to the number of experts by the feature size $M$.
The experts are sparsely activated: for a given input token, each MoE layer's learnable gating function is trained to activate the token-wise best two experts. During inference, the learned gating network dynamically picks the two best experts for each token.
This will results in a model with more capacity while limiting the computation cost.

\subsection{Progressive Expert Expansion}

In the case where only a predefined data distribution exists in the training set, always maintaining a fixed model capacity could be sufficient to fit the pretraining task.
However, when the previously learned language representations cannot account for new data distributions, additional parameters need to be introduced to the network.
Increasing the model capacity via naively expanding the depth/width of networks will also largely increase the computation cost~\citep{zhou2012online,rusu2016progressive,yoon2017lifelong}.
To facilitate the memorization of new corpus without incurring extra computations, we choose to leverage the advantage of MoE: we only increase the number of experts while still sparsely activating two experts for each token.

We need to decide how to expand and initialize new experts and gatings.
We empirically observed that randomly initializing expanded experts and gatings leads to poor performance, potentially due to mismatched gradient directions and magnitudes from new experts/gatings and pretrained dense/attention layers.
Therefore, inspired by the ``Net2WiderNet'' approach~\citep{chen2015net2net}, a better way is to initialize each new expert and gating dimension from pretrained ones, helping both the preservation of old knowledge and the warming-up for the subsequent pretraining.

A vanilla expansion strategy would be to \emph{duplicate} the number of experts in order to fully leverage and inherit all the pretrained knowledge.
However, this will lead to an exponentially increasing model size, which is not scalable.
In our work, we choose to partially expand the number of experts and gating dimensions. We study differ expansion choices, and will show that by expanding a limited number of experts for each data distribution we can achieve competitive performance without further introducing extra model size.
That means, we selectively expand (and only expand) the experts when necessary to accommodate incoming new data distribution that is not covered by the older corpora.
We do not increase the number of dense layers.

\subsection{Expert/Gating Regularization}

The purpose of our expert/gating expansion is to enlarge the model capacity for incoming new data distributions. At this moment, pretrained experts and gatings store the knowledge about previous distributions. Continuous training will still erase these pretrained knowledge and overfit on the new data, which is not desired.
In this section, we propose two approaches to effectively preserve old knowledge.

\paragraph{Implicit Regularization via Distillation from Old Experts/Gatings}

We try to find possible ways to implicitly regularize parameters, including the newly expanded experts, gating dimensions, embeddings, and dense/attention layers.
Inspired by~\cite{li2017learning}, we choose to distill the knowledge from old experts and gatings.
Specifically, denoting the model as $\mathcal{M}$,
we minimize the combination of perplexity loss $\mathcal{L}_{\mathrm{Perp}}$ (for the next-token prediction) and the KL divergence $\mathcal{L}_{\mathrm{KL}}$ of outputs from two models:

{\small
\begin{align}
\mathcal{L} & =\mathcal{L}_{\mathrm{Perp}}+\lambda \mathcal{L}_{\mathrm{KL}} \\
\mathcal{L}_{\mathrm{Perp}} & = - \sum_{\bm{x}_i \in \bm{X}} \log P\left(\bm{x}_{i+1} | \mathcal{M}\left(\bm{x}_{0:i}, \theta_{0:t-1}, \theta_t, \theta_d\right)\right) \\
\mathcal{L}_{\mathrm{KL}} & = - \sum_{\bm{x}_i \in \bm{X}} \mathcal{M}\left(\bm{x}_i, \theta_{0:t-1}, \theta_d\right) \log \left(\mathcal{M}\left(\bm{x}_i, \theta_{0:t-1}, \theta_t, \theta_d\right)\right).
\end{align}
}$\theta_d$ indicates parameters for dense layers that are shared across distributions, $\theta_{0:t-1}$ indicates parameters for old experts and gatings, 
and $\theta_t$ for parameters of newly expanded experts and gating dimensions. $\bm{x}$ is the embedding of the current token and $\bm{X}$ represents the whole corpus of current data distribution.
This auxiliary loss $\mathcal{L}_{\mathrm{KL}}$ will implicitly avoid the model parameters from being updated too far from pretrained ones. It is multiplied with a scaling factor $\lambda$ to control its impact to the original pretraining loss value, and we will study different $\lambda$s.

\paragraph{Explicit Regularization via Partial Experts and Gatings Freezing}
To explicitly preserve pretrained knowledge, an intuitive way is to completely freeze neurons specifically responsible for previous data distributions, and only allow parameters for the current distribution to be updated.
In our method, the dense/attention layers are always being optimized, since they are trained to fit all data distributions.
Newly expanded experts and gating dimensions are also optimized on the new distribution.
Therefore, we only optimize $\mathcal{L}$ regarding $\theta_t, \theta_d$:
\begin{equation}
\theta_t^*,\theta_d^* \leftarrow \underset{\theta_t, \theta_d}{\arg \min }(\mathcal{L}) \label{eq:frozen}
\end{equation}

We will study different freezing strategies: freeze old experts, old gating dimensions, or both.
Old experts and gatings can be regularized (frozen) since we explicitly associate them with each data distribution.
However, since all dense and attention layers are shared across all distributions, we cannot simply freeze their parameters.

\section{Experiment Setup}
\label{sec:exp}

Here, we elaborate our datasets, architecture setting, hyperparameters, pretraining procedure, and evaluation protocol.

\subsection{Training Datasets}

To simulate the distribution-level lifelong pretraining setting,
we build a sequence of billions of tokens that are representative of a wide range of natural language distributions (both English and non-English), based on the GLaM dataset~\citep{du2022glam}.
We collect webpages and Wikipedia pages (with a combination ratio of $81\%~:~19\%$ following~\citep{du2022glam}) as our first distribution, denoted as ``$\mathcal{A}$''.
i18n (``internationalization''), the non-English corpus, will be our second distribution ``$\mathcal{B}$''.
Finally, the conversations from public domain social media~\citep{meena2020} constitutes our third distribution ``$\mathcal{C}$''.
Table~\ref{tab:data} shows the details of our data component sizes and mixture weights.

\begin{table}[h!]
\vspace{-0.5em}
\centering
\small
\captionsetup{font=small}
\caption{Data distributions in our lifelong pretraining set.}
\vspace{-0.5em}
\label{tab:data}
\begin{tabular}{ccc}
\toprule
Distribution & Corpus & Tokens (B) \\ \midrule
$\mathcal{A}$ & \begin{tabular}[c]{@{}c@{}}Wikipedia (19\%)\\ Filtered Webpages (81\%)\end{tabular} & \begin{tabular}[c]{@{}c@{}}3\\ 143\end{tabular} \\ \midrule
$\mathcal{B}$ & i18n & 366 \\ \midrule
$\mathcal{C}$ & Conversations & 174 \\
\bottomrule
\end{tabular}
\vspace{-0.5em}
\end{table}

\textbf{Why these three distributions?}
We design large gaps between these distributions such that catastrophic forgetting issues can be easily observed.
The intuition behind this is that these selections span their contributions to different downstream tasks with less overlap.
The English corpus in the distribution $\mathcal{A}$ will contribute to the downstream QA task~\cite{joshi2017triviaqa}.
The dialogs in $\mathcal{C}$ further diversify the English corpus but contribute less to QA.
In contrast, the non-English materials in distribution $\mathcal{B}$ has zero (or possibly negative) contribution to English-based tasks and will only benefit to translations.
The order of these three distributions is highly related to the study on our downstream tasks: 1) after distribution $\mathcal{A}$, keep pretraining on $\mathcal{B}$ and $\mathcal{C}$ will lead to the forgetting issue on the QA task; 2) after distribution $\mathcal{B}$, keep pretraining on $\mathcal{C}$ will lead to the forgetting issue on the translation task.
We show more studies on influences from these distributions to downstream tasks in our Appendix~\ref{app:dist}.

As we will see in Section~\ref{sec:pretraining} and Figure~\ref{fig:pretrain}, this design explicitly introduces a challenging scenario for our experiments, leading to sharp transitions and a high risk of forgetting issues between corpus distributions. Similar forgetting issues can also be observed in previous works (e.g. Figure 2 in~\cite{hussain2021towards}.

\subsection{Architecture Setting}
\label{sec:exp_setup}

\begin{table*}[h]
\small
    \centering
    \captionsetup{font=small}
    \caption{Sizes and architectures of both our Lifelong-MoE and dense models (Gshard) that we will study in our experiments.
    All trained models share the same learning hyperparameters described in Session~\ref{sec:hyperparam}.
    }
    \label{tab:setup}
    \vspace{-0.5em}
    \begin{tabular}{ccccccccc}
    \toprule 
    $E$ & Type &  $n_{\text{params}}$ &  $n_{\text{act-params}}$ &  $L$ & $M$ &  $H$ &  $n_{\text{heads}}$ &  $d_{\text{head}}$ \\
    \midrule
    4$\sim$16 & MoE & 241$\sim$573M & 145M & 12 & 768 & 3,072 & 12 & 64 \\
    \midrule
    - & Dense & 1.7B & 1.700B & \multirow{2}{*}{24} & \multirow{2}{*}{2,048} & \multirow{2}{*}{8,192} & \multirow{2}{*}{16} & \multirow{2}{*}{128} \\
    16$\sim$32 & MoE & 11$\sim$22B & 1.878B & & & & & \\
    \bottomrule
    \end{tabular} \label{table:model}
    \vspace{-0.5em}
\end{table*}

Table~\ref{tab:setup} shows the hyperparameter settings of different models, ranging from 145 million to 1.878 billion activated parameters.
Here, $E$ is the number of experts (or the dimension of the gating's weight) in each MoE layer, $M$ is the feature/embedding dimension, $H$ is the hidden dimension of the feed-forward layers, $L$ is the number of attention or dense blocks. In addition, $n_{\text{params}}$ is the total number of trainable model parameters, and $n_{\text{act-params}}$ is the number of \emph{activated} model parameters per input token. $n_{\text{heads}}$ is the number of self-attention heads, and $d_{\text{head}}$ is the hidden dimension of each attention head.

\subsection{Hyperparameters}
\label{sec:hyperparam}

We use the same learning hyperparameters for all models and for all data distributions.
More specifically, We use a maximum sequence length of $1024$ tokens in each mini-batch, and pack each input example to have up to 1 million tokens per batch. The dropout rate is set to $0$ since the number of available tokens in the training corpus is much greater than the number of processed tokens during training. Our optimizer is Adafactor~\citep{Shazeer2018AdafactorAL} with first-moment decay $\beta_1=0$, second-moment decay $\beta_2=0.99$ with a $1 - t^{-0.8}$ decay schedule, update clipping threshold of $1.0$, and factored second-moment estimation.
When pretraining on each data distribution, we keep the initial learning rate as $0.01$ for the first 10K training steps, and then decay it with inverse square root schedule $\text{lr} \langle \text{t} \rangle \propto \frac{1}{\sqrt{\text{t}}}$.
We use the SentencePiece~\citep{Kudo2018SentencePieceAS} subword tokenizer with a vocabulary of size of $256$K. During training, we use \textit{float32} for model weights and \textit{bfloat16} for activations. The largest Lifelong-MoE model has 1.878B activated parameters with 40 experts (per expert-layer) and is trained on 128 Cloud TPU-V4 chips.

\begin{figure*}[h!]
\includegraphics[scale=0.65]{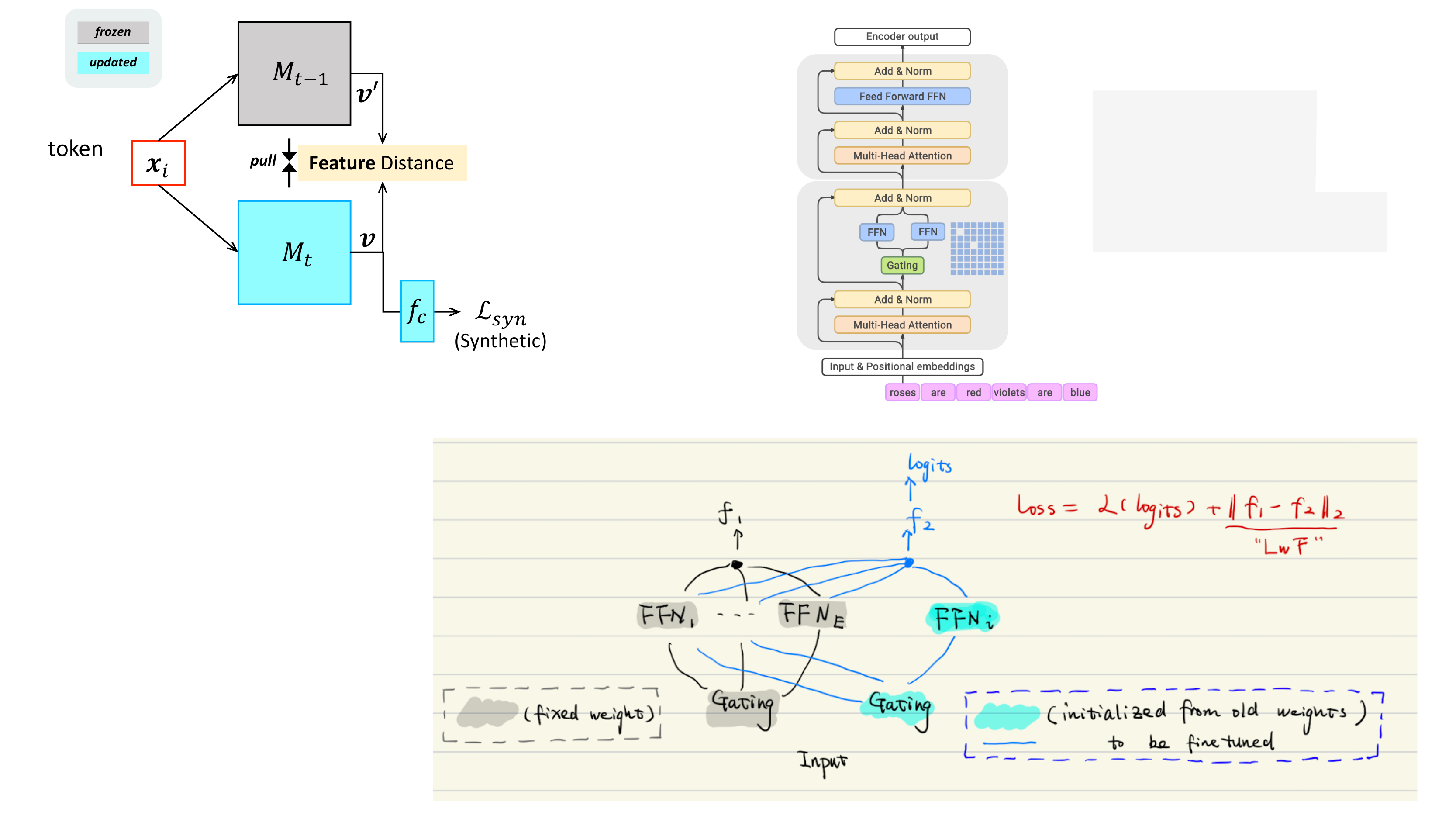}
\centering
\vspace{-0.5em}
\captionsetup{font=small}
\caption{Our method can ameliorate catastrophic forgetting issue in large LMs. Left: next-token accuracy. Right: perplexity. Top/bottom: evaluation on distribution $\mathcal{A}$/$\mathcal{B}$ during lifelong pretraining. We pretrain models on a sequence of data distributions $\mathcal{A} \rightarrow \mathcal{B} \rightarrow \mathcal{C}$. We train on each data distribution for 500K steps. ``0$\sim$500''/``500$\sim$1000'' (K) steps in top/bottom rows represent the pretraining phase on $\mathcal{A}$/$\mathcal{B}$, and subsequent steps stand for forgetting phases (i.e. pretraining on other distributions).}
\label{fig:pretrain}
\vspace{-1em}
\end{figure*}

\subsection{Pretraining Procedure} \label{sec:pretraining}

The pretraining task is to predict the next token in a given sequence with a cross-entropy loss.
To simulate the lifelong pretraining setting, unless explicitly stated, otherwise we will sequentially pretrain models on a distribution streams $\mathcal{A} \rightarrow \mathcal{B} \rightarrow \mathcal{C}$.
On each distribution, the model will first restore the previous checkpoint, and start the pretraining on the new distribution with the same set of hyperparameters.
After pretraining on all three distributions, the model will be evaluated on downstream tasks (described below).
The next-token accuracy and perplexity on all three distributions will be monitored throughout all pretraining phases.

\subsection{Downstream Evaluations}\label{sec:eval_setting}
\paragraph{Protocol.} To clearly demonstrate the effectiveness of Lifelong-MoE models, we mainly focus on evaluating the one-shot and zero-shot decoding tasks suggested by~\citet{radford2019language,NEURIPS2020_gpt3}.
We randomly draw one example from the target task's training set serving as the only demonstration and context. Such a demonstration is concatenated with the evaluation example with two newlines in between, and then fed into the model.

\paragraph{Natural Language Generation Tasks.}
To allow for an apples-to-apples comparison between GShard (densly connected LM)~\citep{lepikhin2020gshard} and our method, we follow the evaluation tasks in \citet{NEURIPS2020_gpt3}.
We mainly study the one-shot decoding task on TriviaQA~\citep{joshi2017triviaqa} and the translation task on WMT16~\citep{bojar2016findings}.
We compare the language sequences decoded by the models to the ground truth in generative tasks.
The performance is measured by the accuracy of exact match (EM) and F1 score, following the standard for each task in~\citet{NEURIPS2020_gpt3}. We use beam search with a width of 4 to generate the sequences.
For WMT16, we calculate the bleu score (bilingual evaluation understudy).

\paragraph{Natural Language Understanding Tasks.} Most language understanding tasks require the model to select one correct answer from multiple options. All binary classification tasks are formulated into the form of selecting among two options (`Yes' or `No'). The prediction is based on the maximum log-likelihood of each option given the context $\log{P(\text{option}|\text{context})}$ normalized by the token length of each option. On a few tasks, such as ReCoRD~\citep{record2018} and COPA~\citep{COPA2012}, the non-normalized loss can yield better results and thus is adopted.
We use the average of the scores reported in all datasets to report the overall few-shot performance of models on NLU tasks.
The F1 scores has been normalized to lie between 0 and 100.

\section{Experiments}

\subsection{Lifelong Pretraining} \label{sec:pretraining}

We first verify that our method can ameliorate the catastrophic  forgetting issue during lifelong pertaining (Figure~\ref{fig:pretrain}).
As we pretrain our lifelong-\glam sequentially on distributions $\mathcal{A} \rightarrow \mathcal{B} \rightarrow \mathcal{C}$, we expect two forgetting phases for $\mathcal{A}$ (when the model is being pretrained on $\mathcal{B}$ and $\mathcal{C}$), and one forgetting phase for $\mathcal{B}$ (when the model is being pretrained on $\mathcal{C}$).
For both next-token accuracy (higher the better) and perplexity (lower the better), we can see huge drops of blue lines at phase transitions. However, our method (red lines) can clearly reduce the drop, retaining the pretrained knowledge from previous distributions.

It is worth noting that this experiment is \emph{to our disadvantage}: the baseline has a constant $10$ experts (per expert layer) throughout all three pretraining phases, whereas we progressively expand the experts ``$4 \rightarrow 7 \rightarrow 10$''. That means, during some phases (e.g. evaluation on $\mathcal{A}$ during 500$\sim$1000K steps), our model with less experts (model capacity) can outperforms the \glam with more experts.

\subsection{Ablation Study}

In this section, we step-by-step study the contributions of expert regularization and expansions to downstream one-shot decoding tasks after the lifelong pretraining.

\paragraph{Output Regularization.}

We first study the choices of different scaling factor ($\lambda$) for our output regularization on a basic \glam model with four experts.
By increasing $\lambda$ from 0, 0.1, to 1, we can improve our F1 score on TriviaQA from 5.93 to 6.96 (row 1$\sim$3 in Table~\ref{table:ablation}).
We also find that $\lambda$ larger than 1 will cause unstable pretraining.

\paragraph{Expert/Gating Freeze.}
An intuitive goal to expand experts is to inherit all pretrained experts into newly expanded ones.
Therefore, starting from 4 experts, our basic expansion strategy is to expand into 8 and 16 experts.

We now study whether to freeze pretrained experts or gating dimentions during training on new distributions.
As shown in Table~\ref{table:ablation} row 4$\sim$7,
freezing either the experts or the gating dimensions are not effective, and only freezing both performs the best.

\paragraph{Partial Expert Expansion.}
Naively duplicating experts and gating dimensions will exponentially increase the model capacity and introduce redundancy.
In our experiments, we study how to achieve comparable performance with reduced experts and gating dimensions.
We explore different expansion ratios, and observe that with ``4$\rightarrow$7$\rightarrow$10'' expert expansion (row 9), we can reduce the model size and achieve even slightly better performance than naive expert duplication.

\begin{table*}[h!]
    \centering
    \captionsetup{font=small}
    \caption{Ablation study of our proposed progressive experts expansion and regularization methods. Results are evaluated on downstream TriviaQA few-shot decoding task after pretraining on $\mathcal{A} \rightarrow \mathcal{B} \rightarrow \mathcal{C}$.}
    \vspace{-0.5em}
    \resizebox{0.61\textwidth}{!}{
\begin{tabular}{ccccc}
\toprule
\# & Expert Expansion & Freeze & Regularization ($\lambda$) & F1 score \\ \midrule
1 & 4$\rightarrow$4$\rightarrow$4 & N/A & 0 & 5.93 \\
2 & 4$\rightarrow$4$\rightarrow$4 & N/A & 0.1 & 5.64 \\
3 & 4$\rightarrow$4$\rightarrow$4 & N/A & 1 & \underline{6.96} \\ \midrule
4 & 4$\rightarrow$8$\rightarrow$16 & N/A & 0 & 6.90 \\
5 & 4$\rightarrow$8$\rightarrow$16 & Experts & 0 & 6.39 \\
6 & 4$\rightarrow$8$\rightarrow$16 & Gatings & 0 & 6.82 \\
7 & 4$\rightarrow$8$\rightarrow$16 & Experts + Gatings & 0 & \underline{6.98} \\ \midrule
8 & 4$\rightarrow$5$\rightarrow$6 & Experts + Gatings & 1 & 5.82 \\
9 & 4$\rightarrow$7$\rightarrow$10 & Experts + Gatings & 1 & \textbf{7.06} \\ \bottomrule
\end{tabular}
}
\label{table:ablation}
\end{table*}

\begin{table*}[h!]
    \centering
    \captionsetup{font=small}
    \caption{Comparison between our Lifelong-MoE with dense GShard~\citep{lepikhin2020gshard}, \glam~\cite{du2022glam}, and classic lifelong learning methods. F1 score is evaluated on TriviaQA. Bleu is evaluated on WMT16.}
    \vspace{-0.5em}
    \resizebox{0.65\textwidth}{!}{
\begin{tabular}{cccccc}
\toprule
Experts & F1 Score & Bleu & Ubuntu & Avg. of 19 NLU Tasks\\ \midrule
Dense + Online L2 Reg. & 12.99 & 5.66 & 27 & 48.65 \\
Dense + Memory Replay & 14.18 & 7.54 & 26 & 48.65 \\
Dense Oracle & 21.25 & 11.14 & 26 & 49.03 \\
\glam & 21.76 & 6.97 & 26 & 50.9 \\ \midrule
Lifelong-MoE (ours) & 20.22 & 19.16 & 27 & 50.26 \\
\bottomrule
\end{tabular}\label{table:1b_comparison}
    }
\end{table*}

\subsection{Lifelong-MoE Mitigates Forgetting Issues in Downstream Tasks} \label{app:lifelong_pretrain}

\begin{table*}[h!]
\centering
\captionsetup{font=small}
    \caption{Decoding results during sequential pretraining on ``$\mathcal{A} \rightarrow \mathcal{B} \rightarrow \mathcal{C}$''.}
    \vspace{-0.5em}
    \label{tab:lifelong_pretraining}
    \resizebox{0.55\textwidth}{!}{
\begin{tabular}{cccc}
\toprule
Method & Phase & TriviaQA F1 & WMT Bleu \\ \midrule
\multirow{3}{*}{Online L2 Reg.} & $\mathcal{A}$ & 25.23 & 2.84 \\
 & $\mathcal{A} \rightarrow \mathcal{B}$ & 17 (-32.6\%) & 20.77 \\
 & $\mathcal{A} \rightarrow \mathcal{B} \rightarrow \mathcal{C}$ & 12.99 (-48.5\%) & 5.66 (-72.7\%) \\ \midrule
\multirow{3}{*}{Memory Replay} & $\mathcal{A}$ & 25.23 & 2.84 \\
 & $\mathcal{A} \rightarrow \mathcal{B}$ & 12.23 (-51.5\%) & 12.34 \\
 & $\mathcal{A} \rightarrow \mathcal{B} \rightarrow \mathcal{C}$ & 14.18 (-43.7\%) & 7.54 (-38.8\%) \\ \midrule
\multirow{3}{*}{Ours} & $\mathcal{A}$ & 33.66 & 4.41 \\
 & $\mathcal{A} \rightarrow \mathcal{B}$ & 26.81 (-20.4\%) & 22.63 \\
 & $\mathcal{A} \rightarrow \mathcal{B} \rightarrow \mathcal{C}$ & 20.22 (-39.9\%) & 19.16 (-15.3\%) \\ \bottomrule
\end{tabular} \label{table:intermediate_results}
    }\vspace{-0.5em}
\end{table*}

Finally, we compare our method with the dense GShard~\citep{lepikhin2020gshard}, \glam~\cite{du2022glam}, and classic lifelong learning methods.

\paragraph{Our Final Large Lifelong-MoE.}
We scale up our final large model of over 1 billion parameters (Table~\ref{table:model}) based on the best expert expansion strategy we found in the last row in Table~\ref{tab:setup}.
We start our lifelong pretraining on distribution $\mathcal{A}$ with 16 experts per expert-layer, and subsequently expand into 28 and 32 for pretraining on distribution $\mathcal{B}$ and $\mathcal{C}$.

\paragraph{Online L2 Regularization.}
The most popular yet simple way of preventing catastrophic forgetting is to regularize the network parameters from deviating too much from its pretrained values using $\ell_2$-regularization~\cite{lin2022continual}, as follows:
\begin{equation}
\begin{aligned}
\minimize_{\bm{W}^{(t)}}\mathcal{L}(\bm{W}^{(t)}; \bm{X}^{(t)}) + \lambda \|\bm{W}^{(t)} - \bm{W}^{(t-1)}\|_2^2
\end{aligned}
\label{eq:l2regular}
\end{equation} 
where $t$ indicates the training step for the current distribution, $\bm{W}^{(t-1)}$ stands for all weights pretrained on the the previous distribution, and $\lambda$ is the regularization scaling factor.
This $\ell_2$-regularization will explicitly enforce the solution $\bm{W}^{(t)}$ to be close to $\bm{W}^{(t-1)}$.
We set $\lambda = 1$ in our experiment.

\paragraph{Memory Replay}

The other important group of lifelong learning methods is based on retraining on previous samples.
Experience Replay (ER)~\citep{Rolnick2019ExperienceRF} is a simple yet effective replay method that stores the previous examples into a growing memory module and periodically sample a small subset of the memory as additional training samples for model training.
We follow the \textit{most competitive setting} in the recent benchmarking work~\citep{lin2022continual}, which sampled one mini-batch of previous data per three mini-batch of current data.
In our experiment, we always keep 25\% historic data when training on a new distribution, i.e., $\mathcal{A} \rightarrow 25\% \mathcal{A} + 75\% \mathcal{B} \rightarrow 25\% (\mathcal{A} + \mathcal{B}) + 75\% \mathcal{C}$.

\paragraph{Joint Pretraining on Multi-distributions}
We can also jointly train a dense LM on our three distributions (with a predefined mixture ratio in~\cite{du2022glam}, as shown in Table~\ref{tab:data}). The LM will see all corpus and serve as the oracle model for comparison. We denote this result as ``Oracle''.

\paragraph{Results}

Our Lifelong-MoE is strong on TriviaQA, WMT16, Ubuntu, and other 19 NLU tasks.
These downstream tasks are associated with our pretraining distributions:
The corpus of TriviaQA is similar to distribution $\mathcal{A}$ (wikipedia + webpages); WMT16 is similar to distribution $\mathcal{B}$ (i18n); Ubuntu and other NLU tasks are similar to distribution $\mathcal{C}$ (conversations).
Therefore, these tasks can faithfully reflect the quality of lifelong pretraining on each distribution.
As shown in Table~\ref{table:1b_comparison}, even comparing with the ``Dense Oracle'', we still achieve better Bleu and NLU scores, with a competitive F1 score on TriviaQA.
Note that
\glam achieves better performance on TriviaQA mainly because it starts with much more experts when training on ``$\mathcal{A}$''.

Moreover, as shown in Table~\ref{tab:lifelong_pretraining}, our method not only demonstrates the best decoding results on TriviaQA and WMT, but also achieves the lowest performance drop (shown in parentheses) when switching to new data distributions.

\section{Conclusion}

In this work, we for the first time aim at solving the data-level lifelong pretraining problem, which considers a stream of online changing distributions in pretraining data resources in NLP tasks, especially for large language models.
Our results demonstrate that, for an MoE architecture, by only introducing extra expert layers, together with appropriate expert/gating regularizations, we can continuously pretrain the MoE on a sequence of data distributions with preserved old knowledge, achieving competitive or even better pretraining quality for downstream tasks.
The expanded experts allocate extra model capacity for new corpus distribution but will not increase computation overhead as the MoE is sparsely activated.
With our method, not only the forgetting issue can be largely mitigated during online pretraining, but each new distribution can be fitting with specific experts. We can achieve state-of-the-art performance on downstream NLU decoding tasks under the lifelong pretraining setting.
We hope our paper could motivate more works and raise more attentions on realistic NLP scenarios during model pretraining, include the distribution shift in pretraining corpus and online pretraining.

\section*{Acknowledgements}
We thank Andrew~Dai for the dataset preparation, Tao~Lei for research ideas on the conditional computation, and Martin~Abadi and Jeff~Dean for insightful discussions and general support.

\bibliography{example_paper}
\bibliographystyle{icml2023}

\newpage
\appendix
\onecolumn


\section{Influence of different distributions on downstream decoding performance.} \label{app:dist}

We also study the influence of different corpus distributions (Table~\ref{tab:data}) on the downstream TriviaQA F1 decoding task.
As shown in Table~\ref{tab:dist_performance}, $\mathcal{A}$ is the most important to TriviaQA, whereas $\mathcal{B}$ will do harms.

\begin{table}[h!]
\centering
    \caption{Influence of different distributions on TriviaQA F1 decoding performance.}
    \label{tab:dist_performance}
\begin{tabular}{cc}
\toprule
Distribution & F1 \\ \midrule
$\mathcal{A}$ & 10.2   \\
$\mathcal{B}$ & 4.64 \\
$\mathcal{C}$ & 7.60 \\
$\mathcal{A} + \mathcal{C}$ & 9.29 \\
\bottomrule
\end{tabular}
\end{table}

\end{document}